\documentclass[a4paper]{esannV2}

\usepackage[utf8]{inputenc} % allow utf-8 input
\usepackage[T1]{fontenc}    % use 8-bit T1 fonts
\usepackage{hyperref}       % hyperlinks
\usepackage{url}            % simple URL typesetting
\usepackage{booktabs}       % professional-quality tables
\usepackage{amsfonts}       % blackboard math symbols
\usepackage{color} %pour mettre de la couleur
\usepackage[super]{nth}
\usepackage{calc}%     needed for the width/height calculations
\usepackage{float}
\usepackage{multirow}
\usepackage{amsmath}
\usepackage{algorithm2e}
\usepackage{algorithmic}
\usepackage{mathtools} % for DeclarePairedDelimiter
\usepackage{bm} %bold math
\usepackage{graphicx,subcaption}

\usepackage{tikz}
\definecolor{mycolor1}{rgb}{1,0.2,0.3}
\definecolor{mycolor2}{rgb}{0.2,0.3,1}
\tikzstyle{tau1} = [mycolor1, dashed]
\tikzstyle{tau2} = [mycolor2, densely dashdotted]
\tikzset{cross/.style={path picture={ 
  \draw[black]
(path picture bounding box.south east) -- (path picture bounding box.north west) (path picture bounding box.south west) -- (path picture bounding box.north east);
}}}

\DeclarePairedDelimiter\abs{\lvert}{\rvert}%
\DeclarePairedDelimiter\norm{\lVert}{\rVert}%
\makeatletter
\let\oldabs\abs
\def\abs{\@ifstar{\oldabs}{\oldabs*}}
\let\oldnorm\norm
\def\norm{\@ifstar{\oldnorm}{\oldnorm*}}
\makeatother

\usepackage{todonotes}

\newcommand{\antoine}[1]{\todo[inline,color=blue!40]{#1 -- Antoine}}  
\newcommand{\isabelle}[1]{\todo[inline,color=orange!40]{#1 -- Isabelle}} 
\newcommand{\benjamin}[1]{\todo[inline,color=red!40]{#1 -- Benjamin}} 

\title{Fast Power system security analysis \\ with Guided Dropout}

\author{\textbf{Benjamin Donnot$^\ddagger$ $^\dagger$}\thanks{Benjamin Donnot corresponding authors: benjamin.donnot@inria.fr}, \textbf{Isabelle Guyon$^\ddagger\bullet$, Marc Schoenauer$^\ddagger$}, \\ \textbf{Antoine Marot$^\dagger$, Patrick Panciatici$^\dagger$} \\
  $\ddagger$ UPSud and Inria TAU, Université Paris-Saclay, France. \\
 $\bullet$ ChaLearn, Berkeley, California.  $\dagger$ RTE France.}

\usepackage[xindy,toc]{glossaries}

\newglossaryentry{line}
{
  name=transmission line,
  description={is a generic term to denote everything that transmit flow in a power grid. Not to be confused with \gls{conn}}
}
\newglossaryentry{lf}
{
  name=load-flow,
  description={is both a model and an algorithm that allow people to compute the flow on each transmission line in a power grid given the injections. It can alternatively be called power-flow}
}
\newglossaryentry{inj}
{
  name=injection,
  description={is an object that inject (or extract) power from a \gls{grid}. Most often, if the value of power injected is positive we will speak of production. If this value is negative, it will be called a load, or consumption}
}
\newglossaryentry{grid}
{
  name=power grid,
  description={aims at transmitting the power from the production to the consumers. Be careful not to confuse it with \gls{ann} or graph}
}
\newglossaryentry{ann}
{
  name=neural network,
  long=artificial neural network,
  description={is a machine learning model. Be careful not to confuse it with \gls{grid} or graph}
}
\newglossaryentry{graph}
{
  name=graph,
  description={is a representation of a \gls{ann}. Not be confused with \gls{ann} or \gls{grid}}
}
\newglossaryentry{conn}
{
  name=connection,
  description={is weight in an \gls{ann}. Not to be confused with \gls{line}}
}
\newglossaryentry{topo}
{
  name=topology,
  description={corresponds to the way the objects (loads, generators, lines etc.) are connected together in an power-grid. Not to be confused with \gls{arch}}
}
\newglossaryentry{arch}
{
  name=architecture,
  description={applies only to neural networks, and corresponds to the way the connections are organized within. Not to be confused with \gls{topo}}
}
\newglossaryentry{reftopo}
{
  name=reference topology,
  description={topology of the power grid before any lines disconnection TODO}
}
\newglossaryentry{skeleton}
{
  name=underlying skeleton,
  description={complete set of connections TODO} 
}
\newglossaryentry{barch}
{
  name=base architecture,
  description={architecture with $\sigma_0$ for reference topology. TODO}
}
\newglossaryentry{mask}
{
  name=mask,
  description={connection pattern TODO}
}
\newglossaryentry{globtopo}
{
  name=global topology,
  description={list/vector of local topologies}
}
\newglossaryentry{loctopo}
{
  name=local topology,
  description={topology of a substation TODO}
}

\definecolor{mycolor1}{rgb}{1,0.2,0.3}
\definecolor{mycolor2}{rgb}{0.2,0.3,1}
\definecolor{mycolor3}{rgb}{0.,0.4,0.}

\begin{document}

\maketitle

%\vspace{-0.5cm}
\begin{abstract}
We propose a new method to efficiently compute load-flows (the steady-state of the power-grid for given productions, consumptions and grid topology), substituting conventional simulators based on differential equation solvers. We use a deep feed-forward neural network trained with load-flows precomputed by simulation. Our architecture permits to train a network on so-called ``n-1'' problems, in which load flows are evaluated for every possible line disconnection, then generalize to ``n-2'' problems without re-training (a clear advantage because of the combinatorial nature of the problem). To that end, we developed a technique bearing similarity with ``dropout'', which we named ``guided dropout''.
\end{abstract}

\section{Background and motivations}

Electricity is a commodity that consumers take for granted and, while governments relaying public opinion (rightfully) request that renewable energies be used increasingly, little is known about what this entails behind the scenes in additional complexity for Transmission Service Operators (TSOs)
to operate the power transmission grid in security. Indeed, renewable energies such as wind and solar power are less predictable than conventional power sources (mainly thermal power plants). 
A power grid is considered to be operated in ``security'' (i.e. in a secure state) if it is outside a zone of ``constraints'', which includes that power flowing in every line does not exceed given limits. 
To that end, it is standard practice to operate the grid in real time with the so-called ``n-1'' criterion: this is a preventive measure requiring that at all times the grid would remain in a safe state even if one component (generators, lines, transformers, etc.) were disconnected. 
Today, the complex task of dispatchers, which are highly trained engineers, consists in analyzing situations
and checking prospectively their effect using sophisticated (but slow) high-end simulators.
As part of a larger project to assist TSOs in their daily operations~\cite{bonnot:hal-01581719}, our goal in this paper is to emulate the power grid with a neural network to provide fast estimations of power flows in all lines given some ``injections'' (electricity productions and consumptions).

\section{The guided dropout method}
Due to the combinatorial nature of changes in power grid topology, it is impractical (and slow) to train one neural network for each topology. Our idea is to train a {\em single} network with architecture variants to capture {\em all} elementary grid topology variants (occurring either by willful or accidental line disconnections) around a reference topology for which all lines are in service.
We train simultaneously on samples obtained with the reference topology and elementary topology changes, limited to one line disconnection (``n-1'' cases), which are encoded in the neural network by activating ``conditional'' hidden units. Regular generalization is evaluated by testing the neural network with additional \{injections, power flows\} input/output pairs for ``n-1'' cases. We also evaluate {\em super-generalization} for ``n-2'' cases, in which a pair of lines is disconnected, by activating simultaneously the corresponding conditional hidden units (though the network was never trained on such cases).

While our work was inspired by ``dropout''~\cite{srivastava2014dropout} and relates to other efforts in the literature to learn to ``sparsify''\cite{bengio2013estimating}
to increase network capacity without increasing computational time 
(used in automatic translation~\cite{shazeer2017outrageously}) and to ``mixed models''
(used e.g. for person identification~\cite{xiao2016learning} or source identification~\cite{ewert2017structured}), we believe that our idea to encode topological changes in the grid in the network architecture is novel and so is the type of application that we address.

\section{Baseline methods}
\begin{figure}[H]    
  \includegraphics[width=\linewidth]{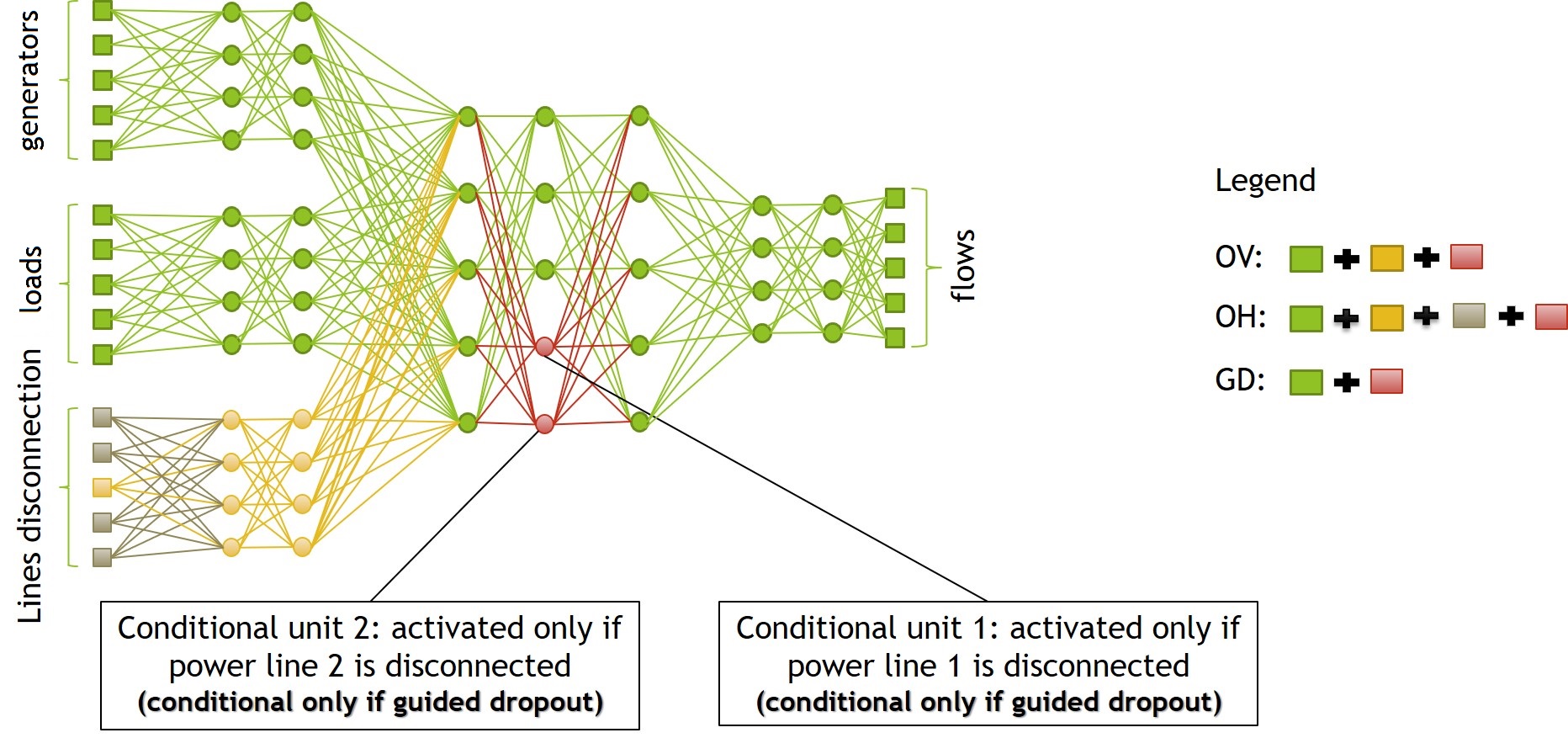}
  \caption{{\bf Neural network architectures being compared.} We overlay all types of architectures under consideration. Lines denoting trainable parameters are not present in all models, as indicated in the legend. Biases not represented. The injection inputs are split into two submodules for productions and consumptions.}
  \label{fig:archi1}
\end{figure}

We compared the performance of our proposed {\bf Guided Dropout (GD)} method with multiple baselines (Figure~\ref{fig:archi1}): \\
$~~~~${\bf One Model}: One neural network is trained for each grid topology. \\
$~~~~${\bf One Var (OV)}: One single input variable encodes which line is disconnected ($0$ for no line disconnected, $1$ for line 1 is disconnected, $2$ for line 2, etc.) \\
$~~~~${\bf One Hot (OH)}: If $n$ is the number of lines in the power grid, $n$ extra binary input variables are added, each one coding for connection/disconnection. \\
$~~~~${\bf DC approximation}: A standard baseline in power systems. This is an approximation of the AC (Alternative Current) non-linear powerflow equations.

{\em One Model} is a brute force approach that does not scale well with the size of the power grid. A power grid with $n$ lines would require training $n(n-1)/2$ neural networks to implement all ``n-2'' cases. {\em One Variable} is our simplest encoding allowing us to train one network for all ``n-1'' cases. However, it does not allow us to generalize to ``n-2'' cases. {\em One Hot} is a reference architecture, which allows us to generalize to ``n-2'' cases.  
The {\em DC approximation} of power flows neglects reactive power and permits to compute relatively fast an approximation of power flows using a matrix inversion, given a detailed physical model of the grid. See our supplemental material for all details on neural network architectures and DC calculations. To conduct a fair comparison towards this DC baseline, we use no more input variables than active power injections. However we should perform even better if we were using additional variables such as voltages as used in AC power flow, not considering them constant as in this DC approximation\footnote{A supplemental material will be available at \href{https://hal.archives-ouvertes.fr/hal-01649938v2}{https://hal.archives-ouvertes.fr/hal-01649938v2} or in the github repository \href{https://github.com/BDonnot/FPSSA-GuidedDropout/blob/master/FPSA_with_GuidedDropout-supplemental_material.pdf}{FPSSA-GuidedDropout}.}.

\section{Experiments}
Our goal in this section is to demonstrate empirically that multiple grid topologies can be modeled with a single neural network with the purpose of providing fast {\bf current flow predictions} (measured in Amps), for given injections and given topologies, to anticipate whether some lines might exceed their thermal limit should a contingency occur. We show a phenomenon that we call {\bf super-generalization}: with the ``guided dropout'' topology encoding, a neural network, trained with ``n-1'' topology cases only, generalizes to ``n-2'' cases. We demonstrate that our approach would be computationally viable on a grid as large as the French Extra High Voltage power grid.

We first conducted systematic experiments on small size benchmark grids from Matpower \cite{Zimmerman11matpowersteadystate}, a library commonly used to test power system algorithms \cite{alsac1974optimal}. We report results on the largest case studied: a 118-node grid with $n=186$ lines. We used all 187 variants of grid topologies with zero or one disconnected line ({\bf ``n-1'' dataset}) and randomly sampled 200 cases of pairs of disconnected lines ({\bf ``n-2'' dataset}), out of $186*185/2$. Training and test data were obtained by generating for each topology considered 10,000 input vectors (including active and reactive injections). To generate semi-realistic data,  we used our knowledge of the French gri, to mimic the spatio-temporal behavior of real data~\cite{bonnot:hal-01581719}. For example, we enforced spatial correlations of productions and consumptions and mimicked production fluctuations, which are sometimes disconnected for maintenance or economical reasons. Target values were then obtained by computing resulting flows in all lines with the AC power flow simulator Hades2,\footnote{A freeware version of Hades2 is available at \url{http://www.rte.itesla-pst.org/}}. This resulted in a ``n-1'' dataset of $1,870,000$ samples (we include in the ``n-1'' dataset samples for the reference topology) and a ``n-2'' dataset of $2,000,000$ samples. We used $50\%$ of the  ``n-1'' dataset for training, $25\%$ for hyper-parameter selection, and $25\%$ for testing. All ``n-2'' data were used solely for testing. 

In all experiments, input and output variables were standardized. We optimized the ``L2 error'' (mean-square error) using the Adam optimizer of Tensorflow.

\begin{figure}
    \begin{center}
    \begin{subfigure}{0.47\textwidth}
        \includegraphics[width=\linewidth]{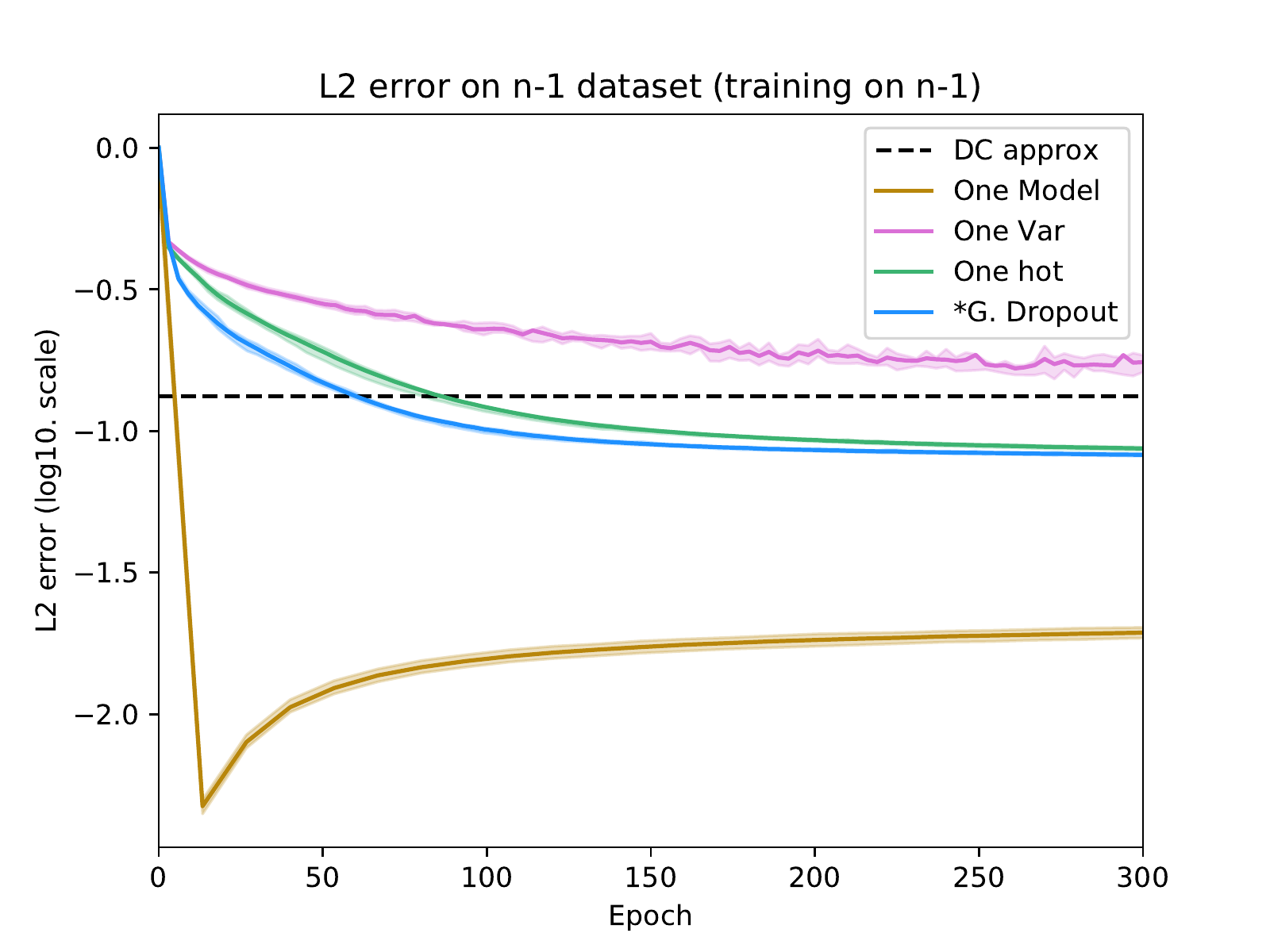}
        \caption{\bf Regular generalization.}
    \end{subfigure}
    \hfill
    \begin{subfigure}{0.47\textwidth}
        \includegraphics[width=\linewidth]{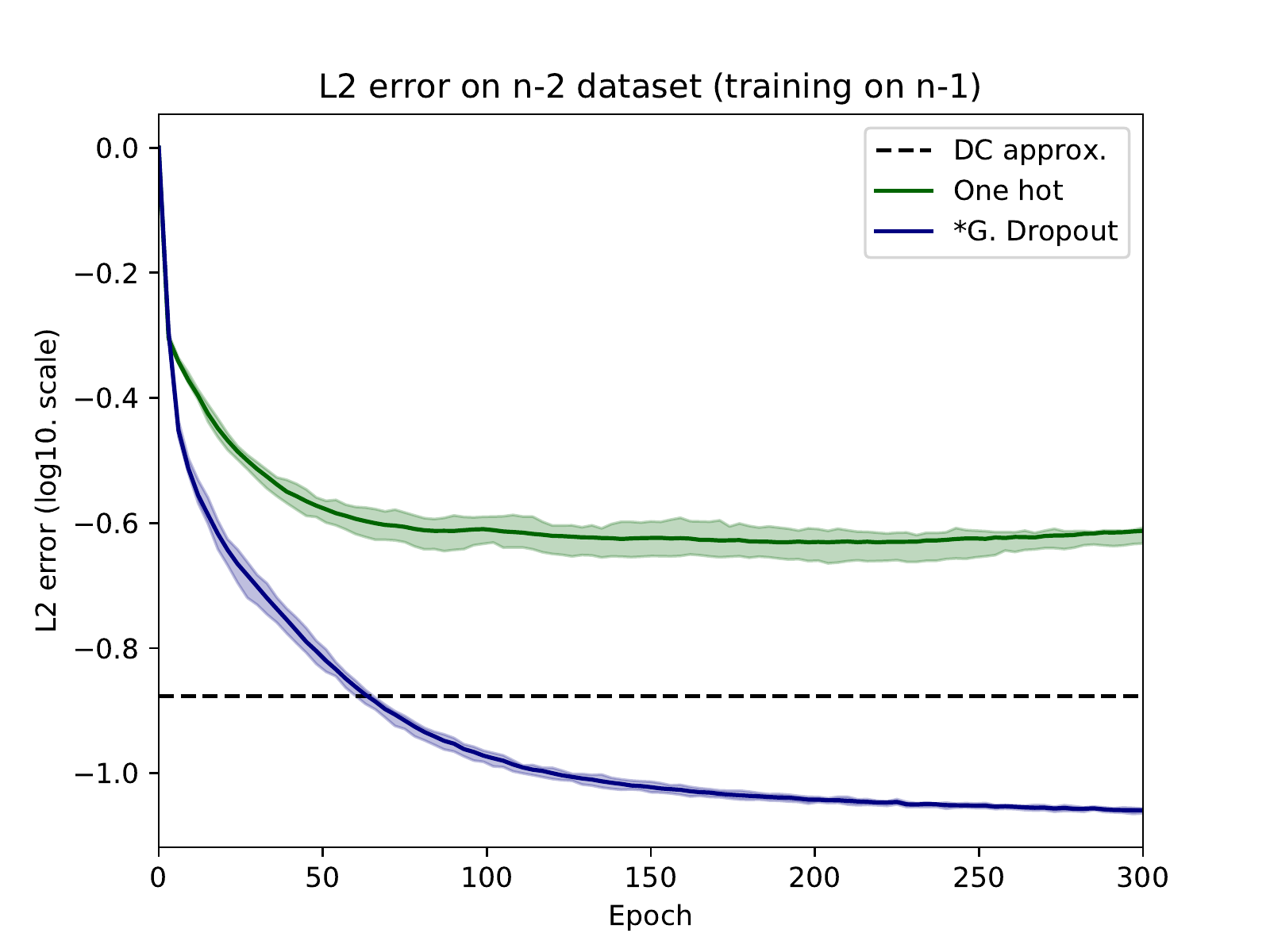}
        \caption{\bf Super-generalization.}
    \end{subfigure}
    \end{center}
   \vspace{-0.3cm} 
    \caption{{\bf Mini grid of 118 buses (nodes).} We show the L2 error in Amperes on a log scale as a function of training epochs. The neural network in both cases is {\bf trained for all ``n-1'' cases} with multiple examples of injections. (a) {\bf Regular generalization.} Test set made of (all) test injections for {\bf ``n-1'' cases}. (b) {\bf Super-generalization.} Test set made of a subset of test injections for {\bf ``n-2'' cases}. Error bars are 25-75\% quantiles over 10 runs having converged.}
    \label{fig:para}
\end{figure} 
 
Figure~\ref{fig:para} shows generalization and super-generalization learning curves for the various methods. For reasons given above, super-generalization can only be achieved by One Hot and Guided Dropout, which explains that Figure~\ref{fig:para}-b has only two curves. The DC approximation is represented as a horizontal dashed line (since it does not involve any training). The test error is represented in a log scale. Hence, it can be seen in Figure~\ref{fig:para}-a that neural networks are very powerful at making load flow predictions since the ``One Model'' approach (yellow curve) outperforms the DC approximation by an order of magnitude.\footnote{We note in the yellow curve some slight over-fitting as evidenced by the test error increase after 10 training epochs, which could be alleviated with early stopping.} However the ``One Model'' approach is impractical for larger grid sizes. The ``One Hot'' approach is significantly worse than both ``Guided Dropout'' and ``DC approximation''. Of all neural network approaches, ``Guided Dropout'' gives the best results, and it beats the DC approximation for ``n-2'' cases (super-generalization). 

The super-generalization capabilities of neural networks trained with ``Guided Dropout'' are obtained by combining in a single network ``shared'' units trained with all available data for many similar (yet different) grid topologies and specialized units activated only for specific topologies. This economy of resources, similar in spirit to weight sharing in convolutional neural networks, performs a kind of regularization.

Obtaining a good performance on new ``unseen'' grid topologies (not available for training) is the biggest practical advantage of ``Guided Dropout'': Acquiring data to train a model for all ``n-2'' cases for the Extra High Voltage French power grid (counting $\simeq 1,400$ nodes, $\simeq 2,700$ lines) would require computing $\simeq 50$ million power flow simulations, which would take almost half a year, given that computing a full AC power flow simulation takes about $300$ ms for RTE current production software. Conversely, RTE stores almost all ``n-1'' cases as part of ``security analyses'' conducted every 5 minutes, so the ``n-1'' dataset is readily available.

To check whether our method scales up to the size of a real power transmission grid, we conducted preliminary experiments using {\em real data} of the French Extra High Voltage power grid with over $1000$ nodes. To that end, we extracted grid state data from September \nth{13} 2011 to October \nth{20} 2014. This represents: $281,543$ grid states, $2735$ lines, $297$ generations units and $1203$ individual loads (accounting for dynamic node splitting, this represented between $1400-1600$ power nodes in that period of time). We did not simulate line disconnections in these preliminary experiments (as we would normally do to apply our method). This is strictly an evaluation of computational performance at {\em run time}. 

We trained a network with the following dimensions (its architecture remains to be optimized): 400 units in the (first) encoder layer, 2735 conditional units in the second (guided dropout) layer, and 400 units in the last (decoder) layer. Training takes of the order of one day.

With this architecture, when the data are loaded in the computer RAM memory, we are able to perform more than $1000$ load-flows per second, which would enables to compute $300$ more load-flows in the same amount of time than current AC power flow simulators.

\section{Conclusions and future work}

Our comparison of various approaches to approximate ``load flows" %(predictions of power flows in electricity transmission grids) 
using neural networks has revealed the superiority of ``Guided Dropout''. This novel method we introduced allows us to train a single neural network to predict power flows for variants of grid topology. Specifically, when trained on all variants with one single disconnected line (``n-1'' scenarios), the network generalizes to variants with TWO disconnected lines (``n-2'' scenarios). Given the combinatorial nature of the problem, this presents significant computational advantages.

%Our method can be viewed as a means of encoding discrete changes in distributions, which cannot be captured with additional neural network inputs. As a result, we can train a network with both continuous variables (regular network inputs) and categorical variables (guided dropout marks), which could find other uses in various applications dealing with heterogeneous data. However, our technique offers unique advantages to generalize to combinations of cases never seen before.

Our target application is to pre-filter serious grid contingencies such as combinations of line disconnections that might lead to equipment damage or service discontinuity. We empirically demonstrated on standard benchmarks of AC power flows that our method compares favorably with several reference baseline methods including the DC approximation, both in terms of predictive accuracy and computational time. 
In daily operations, only ``n-1'' situations are examined by RTE because of the computational cost of AC simulations. ``Guided Dropout'' would allow us to rapidly pre-filter alarming ``n-2'' situations, and then to further investigate them with AC simulation. Preliminary computational scaling simulations performed on the Extra High Voltage French grid indicate the viability of such hybrid approach: A neural network would be $\simeq 300$  times faster than the currently deployed AC power flow simulator.

Given that our new method is {\bf strictly data driven} -- no knowledge used of the physics of the system (e.g. reactance, resistance or admittance of lines) or the topology of grid --, the empirical results presented in this paper are quite encouraging, since the DC approximation DOES make use of such knowledge. This prompted RTE management to commit additional efforts to pursue this line of research. Further work will include incorporating such prior knowledge. The flip side of using a data driven method (compared to the DC approximation) is the need for massive amounts of training data, representative of actual scenarios or situations, in an ever changing environment, and the loss of explainability of the model. The first problem will be addressed by continuously fine tuning/adapting our model. To overcome the second one, we are working on the theoretical foundations of the method. A public version of our code that we will release on Github is under preparation.

\bibliographystyle{plain}
\bibliography{references.bib} 
\end{document}